\documentclass{article}

\usepackage{arxiv}

\usepackage[utf8]{inputenc} 
\usepackage[T1]{fontenc}    
\usepackage{hyperref}       
\usepackage{url}            
\usepackage{booktabs}       
\usepackage{amsfonts}       
\usepackage{nicefrac}       
\usepackage{microtype}      
\usepackage{lipsum,bm}
\usepackage{graphicx}
\graphicspath{ {./images/} }
\newcommand{\argmin}{\operatornamewithlimits{argmin}}
\newcommand{\mb}{\mathbb} 
\newcommand{\vc}{\mathbf} 

\usepackage{amsmath}
\usepackage{algorithm}
\usepackage{algpseudocode}
\newcommand{\mc}{\mathcal}

\title{A robust  kernel machine regression towards biomarker selection in multi-omics datasets of osteoporosis for drug discovery}
\author{
 Md Ashad Alam,  Hui Shen, and Hong-Wen Deng  \\
  Tulane Center for  Biomedical Informatics and Genomics\\
  Division of Biomedical Informatics and Genomics\\ Deming Department of Medicine\\
  Tulane University,  New Orleans, LA 70112, USA\\
  }
\date{}
\begin{document}
\maketitle
\begin{abstract}
Many statistical machine approaches could ultimately highlight novel features of the etiology of complex diseases by analyzing multi-omics data. However, they are sensitive to some deviations in distribution when the observed samples are potentially contaminated with adversarial corrupted outliers (e.g., a fictional data distribution). Likewise, statistical advances lag in supporting comprehensive data-driven analyses of complex multi-omics data integration. We propose a novel non-linear M-estimator-based approach, ``robust kernel machine regression (RobKMR)," to improve the robustness of statistical machine regression and the diversity of fictional data to examine the higher-order composite effect of multi-omics datasets. We address a robust kernel-centered Gram matrix to estimate the model parameters accurately. We also propose a robust score test to assess the marginal and joint Hadamard product of features from multi-omics data. We apply our proposed approach to a multi-omics dataset of osteoporosis (OP)  from  Caucasian females.  Experiments demonstrate that the proposed approach effectively identifies the inter-related risk factors of OP. With solid evidence (p-value $\leq 0.00001$), biological validations, network-based analysis, causal inference, and drug repurposing,  the selected three triplets ((DKK1, SMTN,  DRGX), (MTND5, FASTKD2, CSMD3),  (MTND5, COG3, CSMD3)) are significant biomarkers and directly relate to BMD. Overall, the top three selected genes (DKK1, MTND5, FASTKD2) and one  gene (SIDT1 at p-values $\leq 0.001$) significantly bond with  four drugs- Tacrolimus, Ibandronate, Alendronate, and Bazedoxifene out of $30$ candidates  for drug repurposing in OP. Further, the proposed approach can be applied  to any disease model where multi-omics datasets are available.
\end{abstract}

\section{Introduction}
 Biomedical  technology has accelerated the cycle of multi-omics data discovery for complex disease control and prevention.  Multi-omics data can facilitate our comprehensive understanding of the biological insight into the fundamental mechanism of complex traits and diseases (e.g., osteoporosis (OP)), which represent a significant burden in the global healthcare system \cite{Stewart-00, Clark-15, Rocha-Braz-16, Wang-19}.  These are often a result of the composite interplay between multiple layers of multi-omics data \cite{Hasin-17, Subramanian-20, Tang-21}. An individual-omics technique may detect a subset of biomarkers of a complex disease and thus can only capture changes in a small portion of the biological cascade \cite{Subramanian-20,Zhong-21}. However, a robust integrated risk factors analysis and understanding of comprehensive interactions between various omics data are still rare and challenging. 

In the last decade, several statistical methods have been used to detect gene-gene interactions \cite{ Fan-12, Peng-10, Cheng-17}. Logistic regression, multifactor dimensionality reduction, linkage disequilibrium, entropy-based statistics, and the sequence kernel association test are the  example of such methods  \cite{Wu-13, Dugourd-21}. While most of these methods are based on the unit association of single nucleotide polymorphisms (SNPs), testing the associations between the phenotype and SNPs has limitations. It is not sufficient for interpreting gene-gene interactions \cite{Neafsey-21, Dugourd-21}. However, many researchers have shown that alterations at other levels (i.e., transcriptome, epigenome, proteome, etc.) also play a significant role in complex traits \cite{Venugopalan-21}. Thus, only focusing on genomic data is not sufficient to identify the related risk factors for complex traits. To address this concern, researchers have extended the model-based kernel machine method proposed initially for detecting gene-gene interactions to analyze the interaction of genomic and multiple extra-genomic data to select discriminatory biomarkers \cite{Alam-18c, Ge-15, Li-12, Alam-16a}.

Statistical machine learning approaches (e.g., kernel-based methods) offer beneficial ways to study an extensive collection of genetic variants associated with complex traits. It helps to explore the relationship between genetic markers and a disease state \cite{Wu-13, Ashad-15, Yu-11, Yan-17}. A kernel machine method has been proposed to jointly model the genetic and non-genetic features and their interactions \cite{Ge-15}. While these methods could ultimately highlight novel features of the etiology of complex diseases, they cannot be reliably used for genomic data with multiple extra-genomic data. Recently, researchers have extended these methods to examine the higher-order interaction effect of multimodal ($\leq$ 2 data types) datasets \cite{Alam-21}. However, the major limitation of these methods is their ineffectiveness in presenting outliers or adversarial attacks that often occur in genomics and functional genomics datasets \cite{Du-19, Xu-13}.

\begin{figure}
\begin{center}
\includegraphics[width=7cm, height=3.5cm]{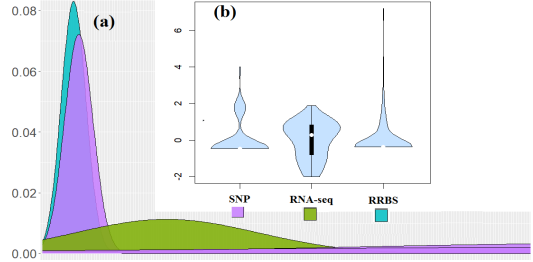}
\caption{ The density (a)  and  volcano plot (b)  of raw multi-omics  data clearly shows that a fictional data distribution and outliers in the dataset.}
\label{fig:dvf}
\end{center}
\end{figure}

To date, almost all genomic analysis methods apply to genome data sets with the assumption that the data sets come from a non-fictional data distribution (a normal distribution only) \cite{Rios-18}. This assumption can lead to an inaccurate inference in real-world genomic data analysis. For example, Figure~\ref{fig:dvf} presents a density (a) and a volcano (b) plot of the genomic (SNP) and extra genomic (RNA-seq and reduced representative bisulfite sequencing (RRBS)) data. This figure clearly shows a fictional data distribution and outliers in the genomic data. However, standard genomic data science approaches are sensitive to such deviations in distribution when the observed samples are potentially contaminated with adversarial corrupted outliers \cite{Alam-19b, Kim-19, Fortino-14, Coretto-18}. Robust learning approaches thus are the critical aspect in achieving true biomedical genomic data integration and are necessary parameters that should not be ignored if our approach to modeling complex diseases is to grow and evolve \cite{Alam-18b}. The novelty is in the use of a sophisticated methodology to examine the interaction and composite effects for network analysis of genomic systems in complex diseases.

\par In biomedical data analysis, robustness is a critical and challenging issue where outliers or adversarial attacks (targeted and untargeted attacks) often occur. Robustness of efficiency means that even if the observations do not match the distributional assumptions properly, the standard error of the statistic will barely be affected. On the other hand, in classical statistical methods, outliers can often cause havoc. To overcome this problem, since the 1960s, many robust methods have been developed, which are less sensitive to the outliers \cite{Huber-09, Hampel-11}. On the other hand, big data-powered machine learning and deep learning approaches are sensitive to fictional data distribution. The advanced methods also need a strong assumption that train and test data came from the same distribution \cite{Rios-18,Ashad-16}. Many robust/adversarial learning techniques have been studied in biomedical data analysis, which is less sensitive to contaminated data and distribution assumptions \cite{Wang-18, Kim-12, Alam-18a}. As a consequence, robust non-linear integrated approaches (e.g.,  robust kernel-based machine)  are an essential feature in the comprehensive analysis of multi-omics biomedical datasets \cite{Nascimento-16, Borgwardt-05, Lanckriet-04}. The robust positive definite kernel-based machine approach can overcome the non-linearity problem and inconsistent results of multi-omics biomedical datasets.

\par Researchers have investigated the issue of robustness for the support vector machine, \cite{Christmann-07,Debruyne-08}, kernel density estimation (Kim and Scott, 2012), kernel canonical correlation, and kernel principal analysis, yet no well-founded robust kernel machine regression method has been proposed. Motivated by these considerations, we introduce a robust kernel machine approach to identify composite effects in multi-omics datasets. Herein, we introduce a novel non-linear M-estimator-based approach, ``robust kernel machine regression", and apply it to identify composite effects in multi-omics data of OP-related traits (e.g., Bone mineral density (BMD)). To achieve robustness, we formulated an empirical optimization problem by combining empirical optimization with the idea of Huber's and Hampel's M-estimation model. The proposed robust kernel machine regression can be computed efficiently via kernelized iteratively re-weighted least squares (KIRWLS) \cite{Alam-21}. 

\par To examine the robustness of our model, we compare the performance of the proposed method with existing methods using synthesized and real OP datasets. OP is a bone disorder that increases bone resorption and (or) decreases bone formation by osteoclasts and osteoblasts. We used three-omics datasets of OP. For more information on the data, we refer the reader to the data construction section \ref{subsec2}. This study explores novel genomes, epigenomes, and transcriptomes that robustly and efficiently identify corresponding factors. To validate the results, we performed network-based analysis, causal inference,  GeneMANIA prediction analysis \footnote{http://genemania.org}, GeneHancer analysis, and drug repurposing. RobKMR shows that the selected three triplets ((DKK1, SMTN,  DRGX), (MTND5, FASTKD2, CSMD3),  (MTND5, COG3, CSMD3)) are significant biomarkers directly related to BMD. In general, the top three selected genes (DKK1, MTND5, FASTKD2) and one gene (SIDT1, at p-values $\leq$ $0.001$) are significantly bonded with four drugs- Tacrolimus, Ibandronate, Alendronate, and Bazedoxifene  out of $30$ candidates  for drug repurposing in OP. The following aspects make this paper highly novel. 

\begin{enumerate}
\item [i.]We proposed a novel non-linear M-estimator-based robust kernel machine regression for improving the robustness of statistical machine regression and the diversity of fictional multi-omics data.
    \item [ii.]	We examined the robust test statistic of the different effects of multi-omics data, including overall interactions and composites.
    \item [iii.]	We analyzed a simulated and a real multi-omics OP dataset to reveal that the proposed method is highly efficient.
    \item [iv.]	To validate the results, we performed a comprehensive GeneMANIA, GeneHancer, pathway, gene-gene-gene network, and causal analysis. The selected biomarkers are significant and directly related to BMD.
    \item [v.] For drug repurposing, we used molecular docking. Using molecular docking,  we observe that the top selected genes significantly bond with candidates lead four drugs- Tacrolimus, Ibandronate, and Bazedoxifene for drug repurposing in OP.
\end{enumerate}

\section{Materials and methods}
\label{Sec:med}
Statistical machine-based approaches utilized in the integrated analysis of multi-omics data that provide a general view of the biological insights of complex diseases and traits \cite{Camps-07, Kung-14}.  These integrated approaches facilitated practical ways to identify interrelated effects, including interactions and composites effects of multi-omics data. Moreover, the critical concern for these  approaches is the fictional data distribution and outliers in the dataset. Such outliers or adversarial attacks often occur in genomics and functional genomics and present a critical challenge for robust data science methods for interactive analyses  \cite{Rios-18}.  Consequently, this paper proposes a robust kernel machine regression via a robust kernel Gram matrix. To that end, different loss functions (e.g., Huber's, Hampel's, and Tukey's bi-weight Cauchy's, Welsch's, and Andrews') is  used in M-estimation(16, 59, 60). Unlike the mean square error loss function, the derivative of these robust loss functions is bounded. We propose an algorithm to estimate a centered robust kernel Gram matrix  using the weight of  robust kernel mean element. The proposed robust kernel machine regression can be computed efficiently via kernelized iteratively re-weighted least squares (KIRWLS) \cite{Alam-18c, Liu-07}.  This method examines the robust test statistic of the different effects, including joint and higher-order interaction and composite effects for identifying risk factors leading to recurrence of the desired complex disease and enhancing disease prediction.

\subsection{Dataset}
\label{subsec2}
We apply the proposed method to our generated multi-omics dataset from an osteoporosis study.  In this paper, we  conducted integrate study of  genome (3,997,535 SNPs which annotated to 25, 442 genes), epigenome (46,690 CpG methylations which annotated to 4,676 genes.),  and transcriptome (22,682 genes expression profiles) data from 57 Caucasian females with high BMD and 51 with low BMD. This dataset (our Louisiana osteoporosis study) is available on request at our lab and some of data has been already deposited in dbGaP (phs001960.v1.p1) \cite{Alam-21}.

\subsection{Robust model setting and estimation}
Robustness to outliers, noisy samples, and heavy-tailed distributions is an essential issue for statistical machine learning approaches, including kernel machine regression. In kernel machine regression, we can  reduce the effect of outliers, noisy samples, or heavy-tailed distribution  using robust $M$-estimation. Consequently, we propose a robust kernel machine regression via robust kernel Gram matrix.   
\subsection{Robust loss function}
\label{Subsec:medk1}
  Different loss functions with bounded derivative  are used in  M (maximum likelihood-type)-estimation. Huber's, Hampel's, and  Tukey's biweight Cauchy's, Welsch's , and Andrews' loss function are common loss functions for the M-estiamtion \cite{Huber-09, Hampel-86, Tukey-77, Wang-20}. The basic notions  of classical and   robust loss function which can be used for  standard  and robust kernel machine approach  are as follows:

 The least-squares loss function is defined as
\begin{eqnarray}
\rho (t)=	 t^2/2.
\end{eqnarray}
This  mean square error loss function is  a  standard and well-known loss function which is used the sum of all the squared differences between the actual value and the estimated value in statistics, statistical machine learning as well as in  data  science. The influence function of this function is unbounded. Hence, least-square estimators are not robust.\\

 The least-absolute  loss function is defined as
\begin{eqnarray}
\rho(t)=	 |t|
\end{eqnarray}
The sum of all the absolute deviations between the estimated value and the true value is used to minimize the error. The least absolute value is an unstable estimator because this loss function $|x|$  is not strictly convex in x. While estimators reduce the influence of large errors than the least-squares estimators, they still have an influence because the influence function has no cut-off point. \\
   
 The  Huber's loss function is defined as
\begin{eqnarray}
\rho (t)=
\begin{cases}
	 t^2/2,\qquad  \qquad 0\leq t\leq c
\\
 ct-c^2/2,\qquad c\leq t, \nonumber
\end{cases}
\end{eqnarray}
where c ($c >0$) is a tuning parameter. This is  a hybrid approach between squared and absolute  error losses functions. It is a parabola around the area of zero and increases linearly at a given level (e.g.,  $|x|>c$).  For almost all situations, the Huber estimator is acceptable; very rarely, it has been found to be inferior to some other loss -functions. However, due to the lack of stability in the gradient values of the function ( e.g., its discontinuous second derivative), this estimator is possibly insufficient.\\

The Hampel's loss function is defined as:
\begin{eqnarray}
\rho (t)=
\begin{cases}
	 t^2/2,\qquad  \qquad\qquad\qquad 0\leq t\le c_1
\\
 c_1t-c_1^2/2,\qquad\qquad  c_1\leq t < c_2
\\
-\frac{c_1}{2(c_3-c_2)}(t-c_3)^2+ \frac{c_1(c_2+c_3-c_1)}{2},\qquad  c_2\leq t < c_3\\
 \frac{c_1(c_2+c_3-c_1)}{2}, \qquad\qquad  c_3\leq t, \nonumber
\end{cases}
\end{eqnarray}
where $c_1 < c_2 < c_3$ the non-negative free parameters  that  allow us to control the degree of suppression.

The Tukey's biweight loss functions is defined as:
\begin{eqnarray}
\rho (t)=
\begin{cases}
	\frac{c^2}{6}( 1-(1-(t/c)^2)^3),\qquad  \qquad 0\leq t\leq c
\\
 \frac{c^2}{6},\qquad c\leq t, \nonumber
\end{cases}
\end{eqnarray}
where $c >0$. Since the loss incurred by large residuals is constant, it is even more insensitive to outliers. But,  like Hubar loss, it reveals quadratic behavior near the origin.  \\

The  Cauchy's  loss function is defined as:
\begin{eqnarray}
\rho (t)=
	\frac{c^2}{2}\rm{log} (1+ (t/c)^2), \nonumber
\end{eqnarray}
where $c$ the non-negative free parameter. Unlike the mean square error and least-absolute  loss functions,  the  Cauchy's  loss function can alleviate the influence of a  considerable noise with a  sample for estimating the residuals. Thus, this loss function  has less dependence on the noise distribution and is more robust to the noise.

The  Welsch's  function is defined as:
 \begin{eqnarray}
\rho (t)=
	\frac{c^2}{2}[1-\rm{exp} (-(t/c)^2)],
\end{eqnarray}
where $c$ the non-negative free parameter. The Welsch's  loss functions try to reduce the effect of significant errors further even suppress the outlier. The windows of this loss function are like a bell. It emphasizes the impact of data with $c$ close to zero and gradually reduces the weights of those further from zero.
 \\
The  Geman-MeClure   is defined as:
 \begin{eqnarray}
\rho (t)=
	\frac{ t^2/2}{1+t^2}. \nonumber
\end{eqnarray}
The Geman-McClure loss function, which behaves almost quadratically for small values and saturates for large ones, is similar to the truncated least squares loss. It produces accurate results and should be preferred over the other ones.

These loss functions hold basic assumptions of the loss functions (i)  non-decreasing,  (ii) weight function exists and is finite,  (iii) influence and weight functions are continuous and bounded, and  (iv) $\varphi(t)$ is Lipschitz  continuous \cite{Kim-12}.

\begin{figure*}[t]
\begin{center}
\includegraphics[width=16cm, height=12cm]{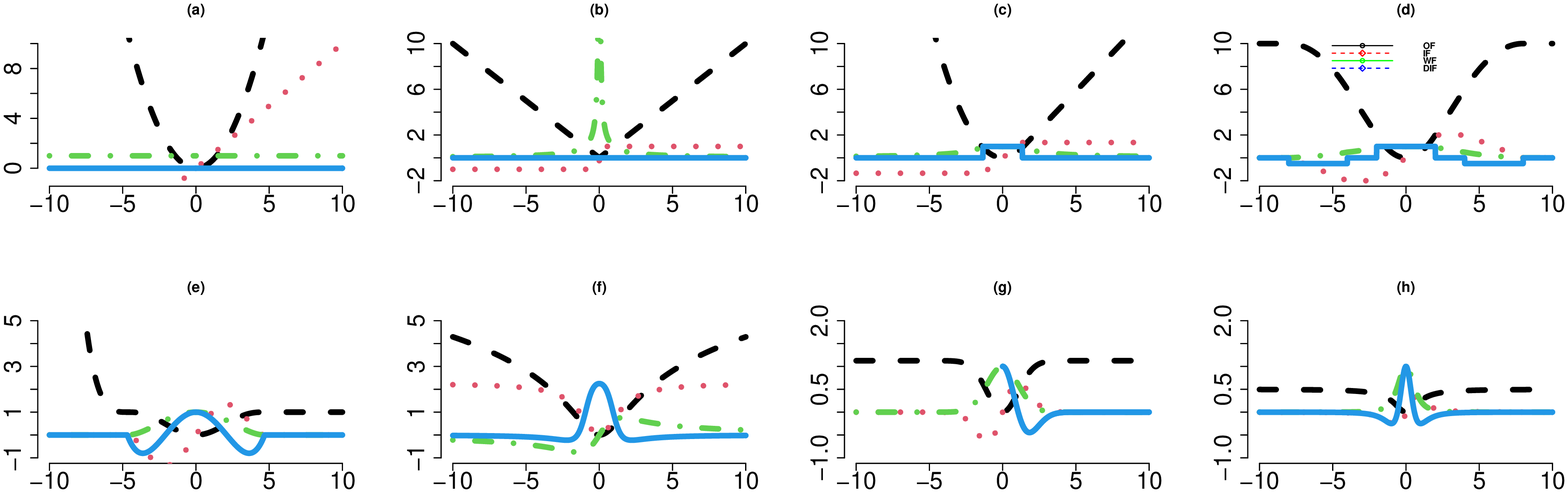}
\caption{Family of  (a) Least-squares, (b) Least-absolute, (c) Huber's, (d) Hampel's,  (e) Tukey's, (f) Cauchy, (g) Welsch, and (h) Geman-MeClure loss functions (OF: original function ($\rho (\cdot))$; IF: influence function, $\psi(\cdot)$; WF: weight function; and DIF: the derivative of influence function).}
\label{Me}
\end{center}
\end{figure*}

\subsection{Robust kernel  Gram matrix}
We propose a robust kernel  Gram matrix using a robust kernel mean element.  The  classical kernel mean element is the solution to the empirical risk optimization problem, which is a least-square  class  of estimators. The method of least squares  is sensitive to the presence of outliers in the data. To reduce the effect of outliers, we can use $M$-estimation for estimating kernel mean element. Kim and Scot (2012) has been proposed  the robust kernel ME  for density estimation  \cite{Kim-12}. The robust kernel ME,  based on  a robust loss function  $\zeta(t)$ on $t \geq 0$, is defined as
\begin{eqnarray}
\label{EKME1}
\widehat{\mc{M}}_R=\argmin_{f\in \mc{H}_X} \sum_{i=1}^n\zeta(\| \Phi(X_i)- f\|_{\mc{H}_X}).
\end{eqnarray}
  Essentially  Eq. (\ref{EKME1})  does not have a closed form solution, but using KIRWLS, the solution of robust kernel mean  is,
  \[\widehat{\mc{M}}_R^{(h)}= \sum_{i=1}^n w_i^{(h-1)}k_X(\cdot, X_i),\]
where $w_i^{(h)}=\frac{\varphi(\|\tilde{\Phi}(X_i) - f^{(h)}\|_{\mc{H}_X})}{\sum_{b=1}^n\varphi(\| \Phi(X_b)- f^{(h)}\|_{\mc{H}_X})}\,, \rm{and} \, \varphi(x)=\frac{\zeta^\prime(x)}{x}.$
the centered robust Gram matrix is 
  \[\tilde{K}_{ij}= \langle \tilde{\Phi}(X_i),\tilde{\Phi}(X_j)\rangle= (\vc{C}\vc{K}_X\vc{C}^T)_{ij},\]
where  $\vc{K}_X = (k_X (X_i, X_j))_{i=1}^n$ is a  Gram matrix, $\vc{1}_n=[1_1, 1_2, \cdots, 1_n]^T$ and $\vc{C}=\vc{I}- \vc{1}_n\vc{w}^T$.

 Given the weights of the robust kernel ME, $\vc{w}=[ w_1, w_2, \cdots, w_n]^T$, of  a set of observations $X_i, \cdots, X_n$,  the points
    \[\tilde{\Phi}(X_i):= \Phi(X_i) - \sum_{a=1}^nw_a \Phi(X_a)\]  are centered.
The algorithm of estimating robust Gram matrix is given in Figure \ref{Robust.K.M}.
\begin{algorithm}
Input: $D=\{\vc{X}_1, \vc{X}_2 ,\ldots \vc{X}_n \}$ in $\mb{R}^m$. The kernel  matrix $\vc{K}$ with kernel $k$ and $\vc{K}_{\vc{X}_i}= k(\cdot, \vc{X}_i)$.   Threshold $TH$, (e.g., $10^{-8}$). The objective function of robust mean element is
\[ M_R=\arg \min_{f\in \mc{H}} J(f),\qquad \rm {where}  \,
J(f)= \frac{1}{n}\sum_{i=1}^n \rho(\|K_{X_i} - f\|_{\mc{H}} )\]
\begin{enumerate}
\item[] Do the following steps until:
\[ \frac{|J(M_R^{(h+1)})- J(M_R^{(h)})|}{J(M_R^{(h)})} < TH,\]
where$ M_R^{(h)}=\sum_{i=1}^nw_i\vc{K}_{\vc{X}_i}, \,
w_i^{(h)}=\frac{\varphi(\|\vc{K}_{\vc{X}_i}- M_R^{(h)}\|_\mc{H})}{\sum_{i=1}^n\varphi(\|\vc{K}_{\vc{X}_i}- M_R^{(h)}\|_\mc{H})}\,, \rm{and} \, \varphi(x)=\frac{\xi^\prime(x)}{x}$
\begin{itemize}
\item[(1)] Set $h=1$ and $w_i^{(0)}=\frac{1}{n}$.
\item[(2)] Solve  $w_i^{(h)}=\frac{ \varphi(\epsilon_i^{[h]})}{\sum_i^n\varphi(\epsilon_i^{[h]})}$ and make a vector $\vc{w}$ for  $i=1, 2, \cdots n$.
\item[(3)] Update the mean element,  $M_R^{(h+1)}=  [\vc{w}^{(h)}]^T\vc{K}$.
\item[(4)] Update error,  $\epsilon^{[h+1]} =(\rm{diag}(\vc{K})- 2[\vc{w}^{(h)}]^T \vc{K}+ [\vc{w}^{(h)}]^T \vc{K}[\vc{w}^{(h)}]^T\vc{1}_n )^{\frac{1}{2}} $.
\item[(5)]  Update $h$ as $h+1$.
\end{itemize}
\end{enumerate}
Output:  the centered robust kernel matrix, $\tilde{\vc{K}}_R= \vc{H}\vc{K}\vc{H}^T$ where $\vc{H}=\vc{I}_n- \vc{1}_n \vc{w}^T$
\vspace*{3mm}
\caption{ The algorithm of estimating centered kernel matrix using robust kernel mean element.}
\label{Robust.K.M}
\end{algorithm}

\subsection{Robust kernel machine regression}
Robustness theory helps us to  understand the behavior of statistical genomic procedures in real-life situations of genomic data without imposing assumptions on the data. While many researchers have been studying the robustness issue in a machine learning setting (e.g.,  support vector machine for classification and regression,  kernel PCA,  kernel CCA, etc.),  a well-founded robust machine learning method has yet to be proposed for  multi-omics \cite{Christmann-04, Christmann-07, Debruyne-08}. 
The standard kernel machine regression associate the output $y_i$ $(i = 1, 2, \cdots, n)$ with $(q-1)$ covariates $
X_i= [X_{i1}, X_{i2},\cdots, X_{i(q-1)}]^T$ and $m$-modal datasets, $\vc{M}_{i}^{(1)}, \vc{M}_{i}^{(2)}, \cdots, \vc{M}_{i}^{(m)}$  of $n$ independent identical distributed (IID) subjects that obey:
\begin{eqnarray}
\label{me1}
y_i= X_i^T\vc{\beta} + f(\vc{M}_{i}^{(1)}, \cdots, \vc{M}_{i}^{(m)}) + \epsilon_i,
\end{eqnarray}
 where  $X_i$ is a  vector of  $q$ covariates including intercept for the $i-$th subject,  $\beta$ is  a  vector of $q$ fixed effects,  $f$ is an unknown function on the product domain, $\mc{M}=\mc{M}^{(1)} \otimes \mc{M}^{(2)}\otimes, \cdots, \otimes \mc{M}^{(m)}$ with  $\vc{M}_{i}^{(\ell)} \in \mc{M}_\ell, \ell= 1, 2, \cdots m$ and the error, $\epsilon_i\sim \rm{NIID}(0, \sigma^2)$. 
By the ANOVA decomposition, $f$ can be decomposed into the main effects, pairwise interactions effects, the interactions effects of the respective dataset, and so on. Similarly, we can also decompose the  functional space, RKHS \cite{Alam-18c, Alam-21}. 

The kernel machine regression is not robust to contaminated data and fictional data distribution.  Several researchers have investigated the issue of robustness for the support vector machine, \cite{Christmann-07, Debruyne-08},  kernel density estimation kernel canonical correlation, kernel principal analysis \cite{Alam-18b}. There are no  well-founded robust   kernel machine regression  method has been proposed yet. Herein, we will introduce a novel non-linear M-estimator-based approach, ``robust kernel machine regression," and apply it to identify composite effects in Genomic and Multiple extra-Genomic Data of disease-related traits. To achieve robustness, we will formulate an empirical optimization problem by combining empirical optimization with the idea of Huber's and Hampel's M-estimation model.

The key goal of this paper is to propose robust  KMR and apply this to three views  (m=3): genome, epigenome, transcriptome, along with  BMD  information of  subjects. To  that end,  assume that  we have   $n$ IID subjects under investigation; $y_i\, (i =1, 2, \cdots n)$ is a quantitative  phenotype for the $i$-th subject.  We also associate the clinical covariates (e.g., age,  weight, and  height) with these five views. Under this setting,  Eq. (\ref{me1}) becomes:
\begin{eqnarray}
\label{me4}
y_i= X_i^T\beta + f(\vc{M}_{i}^{(1)},\vc{M}_{i}^{(2)}, \vc{M}_{i}^{(3)}) + \epsilon_i
\end{eqnarray}
 and 
\begin{multline}
\label{me5}
f(\vc{M}^{(1)}_{i}, \vc{M}^{(2)}_{i}, \vc{M}^{(3)}_{i}) =  h_{\vc{M}^{(1)}}(\vc{M}_{i}^{(1)}) +  \cdots +  \\ h_{\vc{M}^{(1)}\times \vc{M}^{(2)}}(\vc{M}_{i}^{(1)}, \vc{M}_{i}^{(2)})+  \cdots    +\\  h_{\vc{M}^{(1)}\times \vc{M}^{(2)}\times \vc{M}^{(3)}}(\vc{M}_{i}^{(1)}, \vc{M}_{i}^{(2)}, \vc{M}_{i}^{(3)}),
\end{multline}
 Using  the representer theorem \cite{Kimeldorf-71, Schlkof-book} and  the fact that  the  reproducing kernel of  a product of an RKHS is the product of the reproducing kernels \cite{Aron-RKHS}, the expanded  functions of $f$   for arbitrary $\tilde{\vc{M}}^{(\ell)}\in \mc{M}^{(ell)}$, $\ell, \zeta \in \{1, 2, 3\}$  can be written as:
 
\[h_{\vc{M}^{(\ell)}}= \sum_{i=1}^n \alpha^{(1)}_ik^{(1)}(\tilde{\vc{M}}^{(\ell)},\vc{M}_i^{(\ell)}),\]

\[h_{\vc{M}^{(\ell)}\times \vc{M}^{(\zeta)}}= \sum_{i=1}^n \alpha^{(1\times 2)}_ik^{(\ell)}(\tilde{\vc{M}}^{(\ell)},\vc{M}_i^{(\ell)}) 
k^{(\zeta)}(\tilde{\vc{M}}^{(\zeta)},\vc{M}_i^{(\zeta)}),\]
 and 
\[h_{\vc{M}^{(1)}\times \vc{M}^{(2)}\times \vc{M}^{(3)}}=\] \[\sum_{i=1}^n \alpha^{(1\times 2\times 3)}_ik^{(1)}(\tilde{\vc{M}}^{(1)},\vc{M}_i^{(1)})k^{(2)}(\tilde{\vc{M}}^{(2)},\vc{M}_i^{(2)}) k^{(3)}(\tilde{\vc{M}}^{(3)},\vc{M}_i^{(3)}).\]

We can define the robust kernel Gram matrix for  all data set: $\vc{K}^{(1)} = (k^{(1)}(M_i^1, \vc{M}_j^1))_{ij}$, $\vc{K}^{(2)} = (k^{(2)}(M_i^2, \vc{M}_j^2))_{ij}$, $\vc{K}^{(3)} = (k^{(3)}(M_i^3, \vc{M}_j^3))_{ij}$, $\vc{K}^{(1\times 2)} =\vc{K}^{(1)}\odot\vc{K}^{(2)}$,  $\vc{K}^{(1\times 3)} =\vc{K}^{(1)}\odot\vc{K}^{(3)}$, $\vc{K}^{(2\times 3)} =\vc{K}^{(2)}\odot\vc{K}^{(3)}$ and $\vc{K}^{(1\times 2\times 3)} =\vc{K}^{(1)}\odot \vc{K}^{(2)}\odot\vc{K}^{(3)}$, where $\odot$ denotes the  element-wise product of two matrices. Now we have
\[\vc{h}_{\vc{M}^{(\ell)}} = \vc{K}^{(\ell)}\alpha^{(\ell)} \]
\[   \vc{h}_{\vc{M}^{(\ell)}\times \vc{M}^{(\zeta)}} =  \vc{K}^{(\ell\times \zeta)}\alpha^{(\ell\times \zeta)},\]
\[ \vc{h}_{\vc{M}^{(1)}\times \vc{M}^{(2)}\times \vc{M}^{(3)}} =  \vc{K}^{(1\times 2\times 3)}\alpha^{(1\times 2\times\times 3)},\]
where\\
$\alpha^{(\ell)}=[\alpha^{(\ell)}_1, \alpha^{(\ell)}_2, \cdots, \alpha^{(\ell)}_n]^T,$ \,
$\alpha^{(\ell\times \zeta)}=[\alpha^{(\ell\times \zeta)}_1, \alpha^{(\ell\times \zeta)}_2, \cdots, \alpha^{(\ell\times \zeta)}_n]^T,$\, and
$\alpha^{(1\times 2\times 3)}=[\alpha^{(1\times 2\times 3)}_1, \alpha^{(1\times 2\times 3)}_2, \cdots, \alpha^{(1\times 2\times 3)}_n]^T$.
The previous studies established that   the  first-order linear system is equivalent to the normal equation of the linear mixed-effects model  (e.g., \cite{Liu-07, Li-12, Ge-15, Alam-18c,Alam-21}):
\begin{eqnarray}
\label{mee8}
\vc{y}=\vc{X}\beta+\vc{h}_{\vc{M}^{(1)}}+\vc{h}_{\vc{M}^{(2)}}+\vc{h}_{\vc{M}^{(3)}}+ \vc{h}_{\vc{M}^{(1)}\times \vc{M}^{(2)}} + \vc{h}_{\vc{M}^{(1)}\times \vc{M}^{(3)}}\nonumber\\ +\vc{h}_{\vc{M}^{(2)}\times \vc{M}^{(3)}}+ \vc{h}_{\vc{M}^{(1)}\times \vc{M}^{(2)}\times \vc{M}^{(3)}}+\epsilon,
\end{eqnarray}
where $\beta$ is the coefficient vector of fixed effects,
$\vc{h}_{\vc{M}^{(1)}}$, $\vc{h}_{\vc{M}^{(2)}}$, $\vc{h}_{\vc{M}^{(3)}}$,  $\vc{h}_{\vc{M}^{(1)}\times \vc{M}^{(2)}}$,  $\vc{h}_{\vc{M}^{(1)}\times \vc{M}^{(3)}}$,  $\vc{h}_{\vc{M}^{(2)}\times \vc{M}^{(3)}}$ and $\vc{h}_{\vc{M}^{(1)}\times \vc{M}^{(2)}\times \vc{M}^{(3)}}$ are independent random effects with distribution as  $\vc{h}_{\vc{M}^{(1)}}\sim N(0, \tau^{(1)}\vc{K}^{(1)}), \tau^{(1)}= \frac{\sigma^2}{\lambda^{(1)}}$, $\vc{h}_{\vc{M}^{(2)}}\sim N(0, \tau^{(2)}\vc{K}^{(2)}), \tau^{(2)}= \frac{\sigma^2}{\lambda^{(2)}}$, $\vc{h}_{\vc{M}^{(3)}}\sim N(0, \tau^{(3)}\vc{K}^{(3)}), \tau^{(3)}= \frac{\sigma^2}{\lambda^{(3)}}$, $\vc{h}_{M^{(1\times 2)}}\sim N(0, \tau^{(1\times 2)}\vc{K}^{(1\times 2)}), \tau^{(1\times 2)}= \frac{\sigma^2}{\lambda^{(1\times 2)}}$,  $\vc{h}_{M^{(1\times 3)}}\sim N(0, \tau^{(1\times 3)}\vc{K}^{(1\times 3)}), \tau^{(1\times 3)} = \frac{\sigma^2}{\lambda^{(1\times 3)}}$,   $\vc{h}_{M^{(2\times 3)}}\sim N(0, \tau^{(2\times 3)}\vc{K}^{(2\times 3)}), \tau^{(2\times 3)} = \frac{\sigma^2}{\lambda^{(2\times 3)}}$,   $\vc{h}_{M^{(1\times 2\times 3)}}\sim N(0, \tau^{(1\times 2\times 3)}\vc{K}^{(1\times 2\times 3)}), \tau^{(1\times 2\times 3)} = \frac{\sigma^2}{\lambda^{(1\times 2\times 3)}}$. $\epsilon$ is  also an independent random variable with the  distribution $\epsilon\sim N(0, \sigma^2 \vc{I})$, where $\vc{I}$ is an identity matrix.
The effects obtained by minimizing the loss function is the same as the best linear unbiased predictors (BLUPs) of the linear mixed-effects model in Eq. (\ref{mee8}). 

By restricted maximum likelihood (ReML) approach we can  estimate the variance components \cite{Alam-18c, Alam-21}.  The solution gives the coefficients of the  fixed effect, $\beta$, and the  random effect, $\alpha$.
  
\subsection{Statistical testing}
\label{Sec:test}
This section addresses the test statistic of the overall effect, marginal effects,   interaction effects, and composite effects.
\subsubsection{{\bf Overall testing}}
According to our model, the testing overall effect 
 \[H_0:  h_{\vc{M}^{(1)}}= \cdots  =  h_{\vc{M}^{(1)}\times \vc{M}^{(2)}}= \cdots =   \cdots \]
 \[= h_{\vc{M}^{(1)}\times \vc{M}^{(2)}\times \vc{M}^{(3)}\times\vc{M}^{(4)}\times\vc{M}^{(5)}}=0 \]
is equivalent to test the variance components in Eq.(\ref{mee8}),    \[H_0:  \tau^{(1)} =\tau^{(2)} = \tau^{(3)} = \tau^{(1\times 2)} = \tau^{(1\times 3)}=\tau^{2\times 3} =\] \[\cdots=  \tau^{(1\times 2\times 3\times 5\times 5)}=0.\] 
We know that kernel matrices are not block-diagonal. The parameter in variance component analysis is placed on the edge of the parameter space when the null hypothesis is true. Since the asymptotic distribution of a likelihood ratio test (LRT) statistic in favor of the null hypothesis is neither a chi-square distribution nor a  mixture chi-square distribution, we can use a score test statistic on the restricted likelihood \cite{Alam-21, Liu-07}.  The  score test statistic is defined as 
\begin{eqnarray}
\label{overall}
S(\sigma_0^2)= \frac{1}{2\sigma^2_0} (\vc{y}- \vc{X}\hat{\beta})^T \vc{K}(\vc{y}- \vc{X}\hat{\beta}),
\end{eqnarray}
where $\vc{K}= \vc{K}^{(1)}+ \cdots + \vc{K}^{(1\times 2)}+\cdots+ \vc{K}^{(1\times2\times 3)}$ and  $\hat{\beta}$ is the maximum likelihood estimator (MLE) of the  regression coefficients. Under the null model, $\vc{y}= \vc{X}\beta+\epsilon_0$, $\sigma_0^2$ is the variance of $\epsilon_0$ and the  quadratic   function  $S(\sigma_0^2)$ of the variable $\vc{y}$ follows a weighted mixture of the chi-square distribution.  Using the Satterthwaite method, we are able to  approximate the distribution of $S(\sigma_0^2)$ to a scaled chi-square distribution, ($S(\sigma_0^2)\sim \gamma \chi^2_\nu$). For estimating the  the scale parameter $\gamma$ and the degrees of freedom $\nu$,   we use  the method of moments on the mean ($\gamma\nu$) and variance ($2\gamma^2\nu$) of the test statistic that obey:
 $\hat{\gamma}=\frac{\rm{Var}[S(\sigma_0^2)]}{2\rm{E}[S(\sigma_0^2)]}$ and $\hat{\nu}=\frac{2\rm{E}[S(\sigma_0^2)^2]}{\rm{Var}[S(\sigma_0^2)]} $. 
After all, we use the  scaled chi-square distribution $\hat{\gamma}\chi^2_{\hat{\nu}}$ to compute the $p-$ value of the score statistic  $S(\hat{\sigma}_0^2)$.  

  \subsubsection{Testing  composite  effects}
Unlike an interaction hypothesis testing, a composite hypothesis testing is assumed that all lower order effects are statistically significant. Likewise the over all testing, to test the 3rd order composite effect, testing the null hypothesis  $H_0: h_{\vc{M}^{(1)}\times \vc{M}^{(2)}\times \vc{M}^{(3)}\times \vc{M}^{(4)}\times \vc{M}^{(5)}} (\cdot)=0$ is equivalent to testing the variance component: $H_0: \tau^{1\times 2\times 3 \times 4\times 5}=0$. Let $\Sigma= \sigma^2\vc{I} + \tau^{(1)}\vc{K}^{(1)}+ \cdots + \tau^{1\times 2}\vc{K}^{(1\times 2)}+ \cdots + \tau^{ 2\times 3\times 4\times 5}\vc{K}^{(2\times 3\times 4 \times 5)}$, and all $\tau$,  and   $\sigma^2$  are model parameters in favour of  the null model. The  test statistic is defined as:
\begin{eqnarray}
\label{hin}
S_I(\tilde{\theta})= \frac{1}{2\sigma^2_0} \vc{y}^T \vc{B}_I \vc{K}^{(1\times2\times 3 \times 4 \times 5)}\vc{B}_I\vc{y},\end{eqnarray}
 where $\tilde{\theta} = (\sigma^2, \tau^{(1)}, \tau^{(2)},  \tau^{(3)}, \tau^{(1\times 2)}, \tau^{(2\times 3)})$, and  $\vc{B}_I= \Sigma^{-1}- \Sigma^{-1}   \vc{X}(\vc{X}^T  \Sigma^{-1}  \vc{X})^{-1} \vc{X}^T\Sigma^{-1}$ is the projection matrix under the null hypothesis.
 \par We apply  the  Satterthwaite method to  approximate the distribution of   higher order composite test statistic $S_I(\tilde{\theta})$ by a scaled chi-square distribution  with scaled   $\gamma_I$ and degree of freedom $\nu_I$ i.e., $S(\tilde{\theta})\sim \gamma_I \chi^2_{\nu_I}$.  Using MOM, we can compute the  scaled parameter and degree of freedom, $\hat{\gamma}_I=\frac{\rm{Var}[S_I(\tilde{\theta})]}{2\rm{E}[S_I(\tilde{\theta})]}$ and $\hat{\nu}_I=\frac{2\rm{E}[S_I(\tilde{\theta})]}{\rm{Var}[S_I(\tilde{\theta})]}$, receptively.
 In practice, the unknown model parameters are estimated by their respective  ReML estimates favoring the null model.  Finally, the $p-$ value of an  observed higher-order  composite effect test  score statistic  $S_I(\tilde{\theta})$ is  computed by the  scaled chi-square distribution $\hat{\gamma}_I\chi^2_{\hat{\nu_I}}$.

\section{Experiments}
\label{Sec:exp}
In this section, we analyze  the performance of RobKAM on synthetic data and a multi-omics dataset from osteoporosis studies. We compare and evaluate the performance of RobKAM  against the performance of standard baselines kernel machine approach,  principal component SKAT,  and principal component regression (pPCAR).  For the genome  data we use the IBS kernel \cite{Gretton-08, Alam-18c} and for all others dataset we consider  the Gaussian kernel (the median of the pairwise distance as the  bandwidth \cite{Gretton-08, Ashad-15}). For Fisher's scoring algorithm (the ReML algorithm), we follow the parameters setting as in \cite{Alam-21}  to optimize  the proposed  and standard kernel machine approaches.  We conduct a set of initial points in (0, 1) and pick the point which maximized the ReML algorithm to  overcome the potential of  a local minima.

 \subsection{Simulation studies}
In simulation studies, we consider a similar setting as in previous work \cite{Alam-18c, Alam-21} and simulate data in different values of three parameters $(\alpha_1, \alpha_2, \alpha_3)$  for evaluating the performance of the test.   For example,  $\alpha_1= \alpha_2= \alpha_3=0 $  means that all effects have vanished. Then, we analyze the false positive rate for the score test in favor of the overall impact.  For the main effects and 2nd order interaction effects (but no higher-order composite effect), we report evaluating the power of the score test.  We repeat $1000$ simulations for each parameters setting to get consistent results.

To evaluate the power of the composite hypothesis test, we compute the composite score test in different parameter settings.  We state the power of the higher-order composite score test of RobKM and state-of-the-arts methods in Table~\ref{tab:simu1} . In general, we have two observations.  (i)   the test's false positive rate for the higher-order composite effects score test is controlled by fixing the minimal $p-$value threshold to $0.05$ as state-of-the-arts methods  ( the similar observation for the false positive rate) .  (ii)  by considering the power analysis ($\alpha_5\geq 0$), we observed that the proposed method performs better than other methods, and its power exceeds $0.80$ (Table~\ref{tab:simu1}). However, we note that the state-of-the-art methods (pPCAR, fPCAR and SKAT) can significantly   overstate the false positive rates and lose substantial statistical power.

\begin{figure}
\begin{center}
\includegraphics[width=7cm, height=3.5cm]{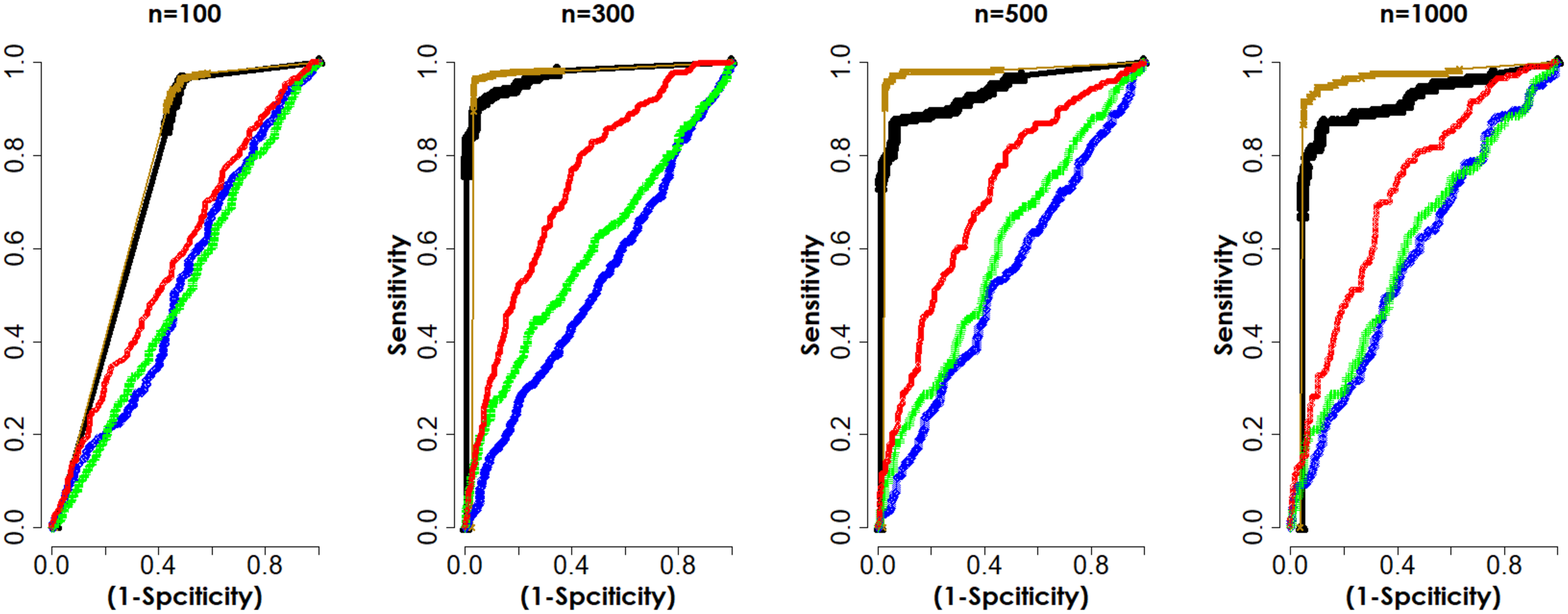}
\caption{The receiver operating characteristics  of the proposed approaches (RobKMA),  the standard   kernel machine approach (stKMA), dimension reduction regression (pPCAR, fPCAR), and sequence kernel association test (SKAT)  with four sample sizes, $n\in \{100, 300, 500, 1000\}$ for  all interrelated parameters ($\alpha_2 = \alpha_3$) values are random but  the linear parameter is fixed to  $\alpha_1 =1$. }
\label{fig:ROC}
\end{center}
\end{figure}

We further visualize the  receiver operating characteristic (ROC)  for three sample sizes, $n\in \{100, 500, 1000\}$.  Figure~\ref{fig:ROC} shows the visual of  the the  ROC  with   related random and the same parameter values  ($\alpha_2 = \alpha_3$)  but the linear parameter is fixed to  $\alpha_1 =1$.  We assign each number with a probability of $0.5$. Also, a random number is uniformly distributed either in a range or at $0$.  By taking a step size $0.0001$,  we plot the sensitivity against (1- specificity) for each $p$-value in the range of $0 - 1$.  In all scenarios, the power gain of RobRKM and the standard kernel machine approach relative to the alternative ones is apparent. This result support that RobRKM has a similar power as the standard kernel Machine approach.

 \begin{table}[!tbp]
\begin{center}
\caption {The power of  higher-order composite  score test  of the proposed approach (RobKAM), and state-of-the-art methods, the  stamdard kernel machine (sKMA),  using dimension reduction regression (pPCAR) and sequence kernel association test (SKAT)}
\scalebox{0.7}[.7]{
\begin{tabular}{|l|cccc|}
\hline
\rm{Parameters}&\multicolumn{4}{c|}{\rm{Simulation. n=300}}\tabularnewline
&\multicolumn{1}{c}{\rm{RobKMR}}&\multicolumn{3}{c|}{\rm{State-of-the-art methods}} \tabularnewline
&\multicolumn{1}{c}{}&\multicolumn{1}{c}{\rm{sKMA}}&\multicolumn{1}{c}{\rm{pPCAR}} &\multicolumn{1}{c|}{\rm{SKAT}} \tabularnewline
\multicolumn{1}{|l|}{($\alpha_1$, $\alpha_2$, $\alpha_3$)}&&&&\tabularnewline
\hline
(0.1, 0, 0)&$0.061$&$0.057$&$0.051$&$0.044$\tabularnewline
(0, 0,  0.1)&$0.852$&$0.842$&$0.056$&$0.044$\tabularnewline
(0, 0, 0.5)&$0.863$&$0.846$&$0.051$&$0.042$\tabularnewline
(0, 0, 1)&$0.812$&$0.824$&$0.052$&$0.041$\tabularnewline
(1, 1, 0.1)&$0.840$&$0.851$&$0.048$&$0.044$\tabularnewline
(1, 1, 0.5)&$0.836$&$0.862$&$0.051$&$0.041$\tabularnewline
(1,  1, 1)&$0.844$&$0.859$&$0.049$&$0.042$\tabularnewline
\hline
\end{tabular}
}
\label{tab:simu1}
\end{center}
\end{table}

\subsection{ Osteoporosis  data analysis}
We apply the proposed method to our generated multi-omics dataset from   Louisiana osteoporosis studies data (as stated in Section $3.1.2$) \cite{Yu-20, Qiu-20}.  The dataset is available at our lab, and some of the data has already been deposited in dbGaP (phs001960.v1.p1).  This dataset included genome (3,997,535 SNPs which annotated to 25, 442 genes), epigenome (46,690 CpG methylations which annotated to 4,676 genes.),  and transcriptome (22,682 genes expression profiles), from 57 Caucasian females with high BMD and $51$ with low BMD.

\begin{figure}
\begin{center}
\includegraphics[width=10cm,height=5cm]{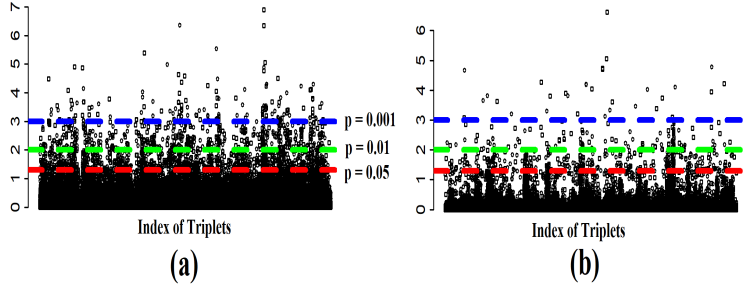}
\caption{  Manhattan plots of negative log of the p-values  $-log_{10}(p)$  in the y-axis against  triplets in the x-axis  based on all significant overall score tests of the  standard (a) and the proposed robust (b) methods.}
\label{fig:p-values}
\end{center}
\end{figure}

We fix each feature (gene) of the genome, epigenome, and transcriptome data as an individual testing unit.   We reduced the dimensionality of all these datasets to make  doable. To that end, we apply different methods (the t-test, canonical correlation analysis based gene shaving (CCAOut),  kernel canonical correlation analysis based gene shaving  (KCCAOut), and Linear Models for Microarray-based gene shaving (LIMMA)) methods for three datasets \cite{Rocha-Braz-16, Alam-21}. Finally,  for our experiment, we consider $100$, $53$, and $61$  genes for the genome,  epigenome, and transcriptome data, respectively. Hence,  we have $ 323300$ ($100 \times 53\times 61)$   triplets to test the overall and higher-order composite effect. The overall test of the RobKMR approach and stKMR approach provides us with  $102973$ and $ 75525 $ significant triples ($p \leq 0.05$), respectively. In Figure~ \ref{fig:p-values}, we exhibit the plot of $-\rm{log}_{10}(p)$ for the both RobKMR and stKMA approaches. The vertical solid, dotted, and double dotted lines correspond to the p-values of $0.05$, $0.01$, and $0.001$, respectively. Table \ref{tbl:gnu} shows the number of significant genes selected at different p-values by the stKMA and the RobKMR methods.   This table clearly shows that the proposed method (RobKMR) can identify a small set of genes at all p-values. 

\begin{figure}
\begin{center}
\includegraphics[width=8cm,height=5cm]{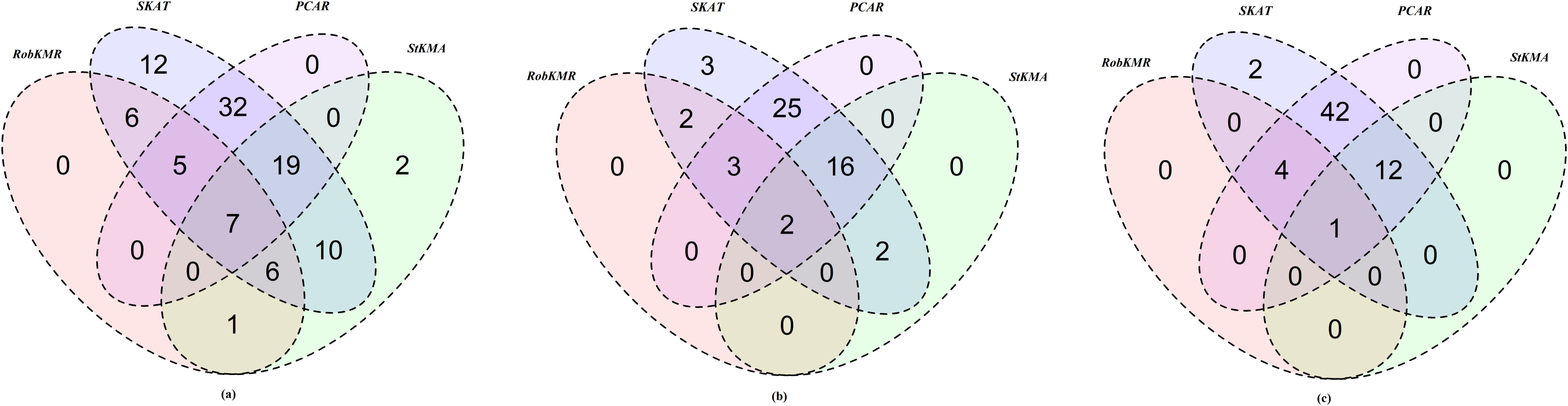}
\caption{ Venn diagrams of the selected genes of  each three datasets: (a) genome, (b) epigenome, and (c) transcriptome using  the proposed and other three methods.}
\label{fig:VD}
\end{center}
\end{figure}

Table~ \ref{tab:tgrg} presents the ReML estimates of all parameters, $\sigma^2$, $\tau^{(1)}$,  $\tau^{(2)}$, $\tau^{(3)}$,  $\tau^{(1\times 2)}$,  $\tau^{(1\times 3)}$,  $\tau^{(2\times 3)}$,  $\tau^{(1\times 2\times 3)}$ and  the $p$-values for both the proposed and SKAT methods for  each of the  $10$ triplets.  By the proposed method, these  $10$ triplets  were identified to have significant interactions at a level of  $p \leq  0.0000624$. At this  $p$-value, we observe that the   unique  $8$  genes ({\bf DKK1,  MTND5, WNT3,  MPP7,  ANAPC1, FUBP3, YWHAE}, {\bf LRP5})   $5$ genes ({\bf SMTN, FASTKD2, COG3, DNMBP}, {\bf NMBP }), and  $4$ genes ({\bf DRGX, CSMD3, SOX1}, {\bf USP17L1}),  are selected from genome, epigenome and transcriptome data, respectively. For the stKMA,  at P-value $0.00001$ and $0.000001$,  the selected genes for three datasets are   {\bf GPLD1,     LINC00461,  PANK1,     SOST,   ATP2B2,,  KNDC1,  NUP214,  CAPN3   PRPF38B} and  ({\bf   LINC00461,      SOST,  KNDC1,    CAPN3   PRPF38B}, respectively.  On the other hand, for the RobKMR, at same p-values, the selected genes are   {\bf DDK1,     MTND5, COG3,   FASTKD2, SMTN,  CSMD3,  DRGX} and  {\bf DDK1,     MTND5,   FASTKD2, SMTN,  CSMD3,  DRGX}, respectively.  At these low  p-values,  we observe the different set of genes for the stKMA and RobKMR methods. In addition,  a list of genes (genome, epigenome, and transcriptome data) for the RobKMR  is  tabulated in Table~\ref{tbl:gnu} at a p-value of $0.001$.  Figure  \ref{fig:VD} presents  Venn diagrams of the selected genes of  each three datasets (genome, epigenome, and transcriptome) using  the proposed and other three methods. This figure also shows the proposed method is  able to identify a least set of genes than other methods.

\begin{table*}
\begin{center}
 \caption {The selected significant genes   using the proposed method and  state-of-the-art methods. The $p-$value was set to be $0.0000624$ for top 10 triples (OV: Overall effect and HOC: Higher-order composite effect)}
\scalebox{0.6}[0.6]{
\begin{tabular}{lcccccccccccccccc}
\hline
&&&\multicolumn{10}{c}{\rm{RobKMR}}&\multicolumn{3}{c}{\rm{State-of-the-Art Methods}}\tabularnewline
\multicolumn{1}{l}{Genome}&\multicolumn{1}{c}{Epigenome}&\multicolumn{1}{c}{Transcriptome}&\multicolumn{1}{c}{$\sigma^2$}&\multicolumn{1}{c}{$\tau^{(1)}$}&\multicolumn{1}{c}{$\tau^{(2)}$}&\multicolumn{1}{c}{$\tau^{(3)}$}&\multicolumn{1}{c}{$\tau^{1\times 2}$}&\multicolumn{1}{c}{$\tau^{1\times 3}$}&\multicolumn{1}{c}{$\tau^{2\times 3}$}&\multicolumn{1}{c}{$\tau^{1\times 2\times 3}$} &\multicolumn{1}{c}{\rm{OV}} &\multicolumn{1}{c}{\rm{HOC}} &\multicolumn{1}{c}{\rm{StKMA}}&\multicolumn{1}{c}{\rm{SKAT}} &\multicolumn{1}{c}{\rm{pPCAR}} \tabularnewline\hline
  {\bf DKK1}& {\bf SMTN}&  {\bf DRGX}&  $0.0179$&$4e-04$&$ 0.0019 $&$0.0073$&$ 0.0044$&$ 0.0001$&$ 0.0383$&$ 0.0000$&$ 0.0911$&$2.71e-10$ &$1.0000 $ &$0.00012 $ &$0.0393$\tabularnewline
    {\bf MTND5}&   {\bf FASTKD2}&     {\bf CSMD3}&$0.0343$&$ 0e+00$&$ 0.0001$&$ 0.0001$&$ 0.0001$&$ 0.0001$&$ 0.0000$&$ 0.0100$&$ 0.0433$&$ 2.45e-07$ &$0.0004$ &$0.9853$ &$0.2354 $   \tabularnewline
    {\bf MTND5}&   {\bf  COG3}&   {\bf CSMD3}&$ 0.0342$&$ 5e-04$&$ 0.0010$&$ 0.0009$&$ 0.0010$&$ 0.0012$&$ 0.0000$&$ 0.0100$&$ 0.0167$&$ 8.67e-06$ &$1.0000 $ &$ 0.0003$  &$0.1903$  \tabularnewline
    {\bf WNT3}&    {\bf DNMBP}& {\bf SOX1}&$ 0.0343$&$ 1e-03$&$ 0.0010$&$ 0.0010$&$ 0.0010$&$ 0.0010$&$ 0.0000$&$ 0.0000$&$ 0.0000$&$1.63e-05$ &$1.0000 $ &$0.0124$  &$0.3547$   \tabularnewline 
    {\bf MPP7}&    {\bf DNMBP}&      {\bf DRGX}&$0.0340$&$1e-03$&$0.0010$&$0.0010$&$ 0.0005$&$ 0.0010$&$ 0.0000$&$ 0.0000$&$ 0.0000$&$ 1.96e-05$ &$1.0000 $ &$0.8562 $  &$0.1529$   \tabularnewline
    {\bf MPP7}&    {\bf DNMBP}&     {\bf SOX1}&$0.0343$&$ 1e-03$&$ 0.0010$&$ 0.0010$&$ 0.0009$&$ 0.0010 $&$0.0000$&$ 0.0000$&$ 0.0000$&$ 2.10e-05$ &$1.0000 $ &$0.6842 $ &$0.0265$   \tabularnewline
    {\bf ANAPC1}&     {\bf DNMBP}&   {\bf USP17L1}&  $0.0342$&$ 1e-03$&$0.0010$&$ 0.0012$&$ 0.0010$&$ 0.0000$&$ 0.0004$&$0.0100$&$0.0000$ &$5.31e-05$& $1.0000 $ &$1.0000 $ & $ 0.2564$   \tabularnewline
      {\bf FUBP3 }& {\bf NMBP}&        {\bf SOX1}&$ 0.0343$&$ 1e-03$&$ 0.0010$&$ 0.0010$&$ 0.0010$&$ 0.0010 $&$0.0000$&$ 0.0000$&$ 0.0001 $&$5.97e-05$  &$1.0000$ &$0.0125 $ &$0.3602 $   \tabularnewline
  {\bf YWHAE }&    {\bf DNMBP}& {\bf SOX1}&$ 0.0343$&$ 1e-03$&$ 0.0010$&$ 0.0010$&$ 0.0010$&$ 0.0010$&$ 0.0000$&$ 0.0000$&$ 0.0001$&$6.24e-06$ &$0.0235 $ &$0.0285$  &$0.4369$   \tabularnewline
    {\bf LRP5}&   {\bf  DNMBP}&      {\bf DRGX}&$ 0.0341$&$ 8e-04$&$ 0.0010$&$ 0.0011$&$ 0.0000$&$ 0.0010$&$ 0.0000$&$ 0.0032$&$0.0001$&$ 6.54e-05$ &$0.0091 $ &$0.0002 $ &$0.3602$  \tabularnewline  \hline 
      \end{tabular}
}
\label{tab:tgrg}
\end{center}
\end{table*}

\begin{table}
\begin{center}
 \caption {The number of significant  genes  selected at differect p-values by the stKMA and RobKMR methods}
\scalebox{0.6}[1]{
\label{tbl:gnu}
 \begin{tabular}{l|ccc|ccc} \hline
&\multicolumn{3}{c}{KMRHCE}&\multicolumn{3}{c}{ RKMRHCE }\\ \cline{2-7}
P-values&Genome& Epigenome & Transcriptome &Genome& Epigenome & Transcriptome   \\ \hline
$0.05$& $98$&$45$ &$36$ & $90$&$40$ &$44$ \tabularnewline\hline
$0.01$& $90$&$30$ &$21$ & $58$&$23$ &$26$ \tabularnewline\hline
$0.001$& $45$&$20$ &$13$ & $25$&$7$ &$5$ \tabularnewline\hline
$0.0001$& $18$&$6$ &$4$ & $11$&$4$ &$4$ \tabularnewline\hline
$0.00001$& $4$&$3$ &$2$ & $2$&$3$ &$2$ \tabularnewline\hline
$0.000001$& $2$&$1$ &$2$ & $2$ &$2$ & $2$ \tabularnewline\hline
\end{tabular}
}
\end{center}
 \end{table}

\begin{table}[!tbp]
\begin{center}
 \caption {The selected  significant genes by the RobKMA method at p-value $= 0.001$. The highlighted  genes identified  by   RobKMA  only.}
\scalebox{0.45}[.7]{
\begin{tabular}{lcccccccccccccccc}
\tabularnewline\cline{1-10}
\rm{Method}&&&& \rm{Genome}&&\tabularnewline\cline{1-10}
& ANAPC1& {\bf  CLCN7}&CPN1&CSF1&{\bf DKK1}&FUBP3&GPATCH1&GPLD1
&{\bf INSIG2}\\
\rm{Genome}&JAG1&{\bf LIN7C}&LRP5& {\bf MBL2}& MPP7& MTND5P19& NIPAL1&{\bf NTAN1}&{\bf RGCC}\\&RUNX2&{\bf SIDT1}&{\bf SLC25A13}&SOST&SOX6&WNT3&{\bf YWHAE}\tabularnewline\cline{2-10}
\tabularnewline
\hline
\rm{Epigenome}&{\bf COG3}& {\bf DNER }& {\bf DNMBP}& FASTKD2& PLD5& { \bf RBM38} & { \bf SMTN}& \tabularnewline\cline{2-10}
\hline
\rm{Transcriptome}& CSMD3& {\bf DRGX}& {\bf SOX1}& {\bf UCKL1}& {\bf USP17L1}&& \tabularnewline\cline{2-10}
\hline
\end{tabular}
}
\label{tab:grg}
\end{center}
\end{table}

\par
\begin{figure}
\begin{center}
\includegraphics[width=8cm, height=8cm]{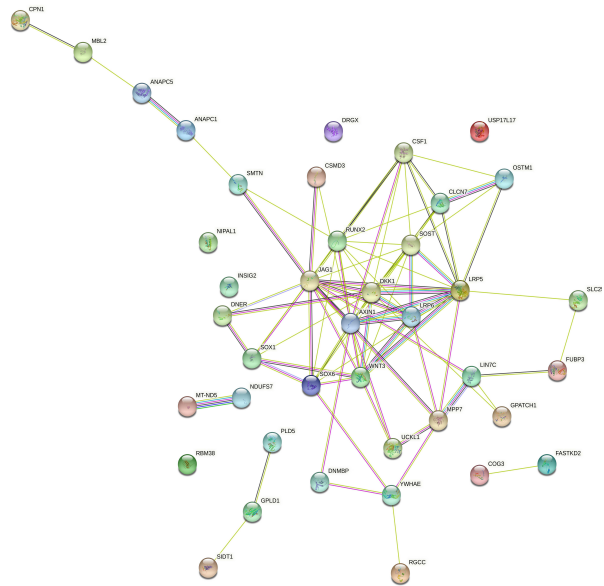}
 \caption{The  functional protein association networks of the selected genes of  genome, epigenome, and  transcriptome datasets by the proposed method (at p-values $leq 0.001$).} 
\label{fig:ggg3d}
\end{center}
\end{figure}

  We also construct functional protein association networks for available enrichment analysis using STRING  \footnote{https://string-db.org/}. Figure~ \ref{fig:ggg3d} shows the gene-gene networks based on the protein interactions among the selected genes of all three datasets. ($37 = 25+07+05$, at p-values $\leq$ 0.001) by the RobKMR. The network analysis demonstrates that the number of nodes, edges, expected edges, average node degree, clustering coefficient, protein-protein interaction enrichment p-values are $36$, $62$, $32$,  $3.44$,  $0.518$ and $0.0000012$, respectively. In addition, this network has  $2$ GO-terms, $525$ publications,  $5$ pathways, and $6$ diseases significantly enriched. In this figure, the color saturation of the edges represents the confidence score of a functional association. Therefore, this network analysis confirms that the selected genes Have significant interactions than expected. It also indicates that the most selected genes may function collaboratively.

\par We access the biomedical and genomic information using the Human gene database (GeneCards)  \footnote{ https://www.genecards.org/} and the GeneMANIA \footnote{ https://genemania.org/ }  to confirm the biological roles of the selected genes \cite{Fishilevich-17}.   As we know,  the critical goal of the GeneCards database is the unequivocal identification of enhancer elements and uncovering their connections to genes for understanding gene regulation and molecular pathways. On the other hand,  GeneMANIA finds other genes related to a gene or a set of input genes using an extensive set of functional association data. This organized data includes protein and genetic interactions, pathways, co-expression, co-localization, and protein domain similarity.
We used the GeneCards database to provide insight into the gene regulatory elements (promoters and enhancers) for $7$ selected genes (DKK1,  MTND5, COG3, FASTKD2,  SMTN, CSMD3, and DRGX) at p-value $\leq 0.00001$ by the RobKMR. Table~ \ref{tbl:ostdatagene}   shows GeneHancer identifier, GeneHancer score, gene association score, total score, major-related diseases, and PubMed database. This table shows that the selected genes have a remarkable GeneHancer score, gene association score, total score, and literature review in the past studies. According to the disease annotation, the selected  7 genes are highly associated with complex diseases, including the higher risk of developing osteoporosis.

\par
\begin{figure}
\begin{center}
\includegraphics[width=8cm, height=8cm]{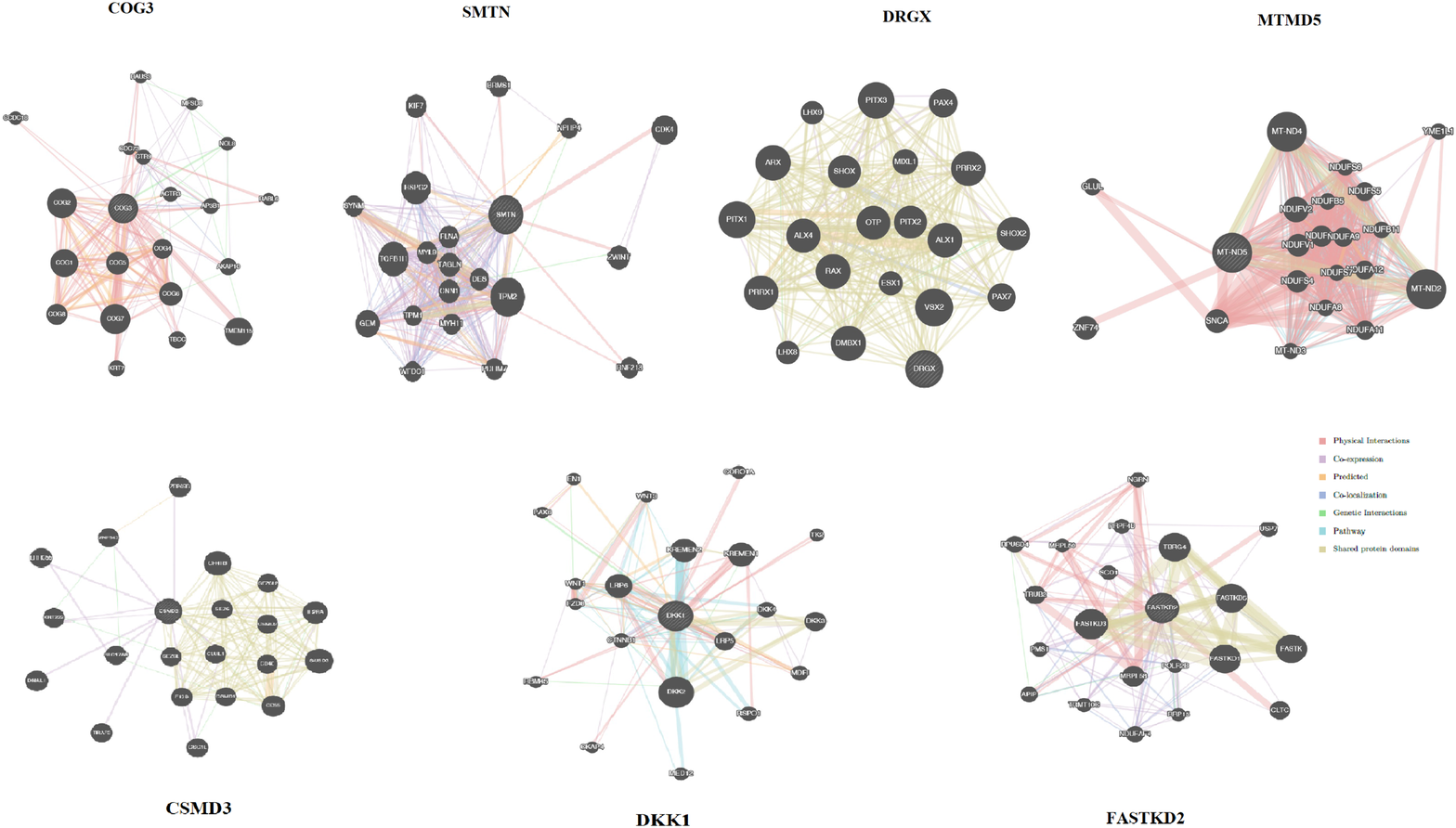}
 \caption{The network of  predicted related genes of  each unique selected  genes by the proposed method  at p-value $\leq$ 0.00001.} 
\label{fig:mg}
\end{center}
\end{figure}

 As a comparison between the stKMA and RobKMR,  stKMA  extract  $78 (45+20+13)$ significant genes, while the  RobKMR  extract  $37 (25+7+5)$ significant genes at a  p-values $\leq 0.00001$. The RobKMR extracts $20$ unique genes out of 37 (i. e., 17 genes are common for both methods). We intend to show the performance that the RobKMR could find undiscovered genes and could exclude not significant genes (in the meaning of robustness). To that end, we also conduct the network analysis of genes related to an input gene or a set of input genes using an extensive set of functional association data with the GeneMANIA.  Table~ \ref{tbl:norm}  presents the network analysis of each gene along with the number of edges,  average node degree, average local clustering coefficient, expected number of edges, and interaction enrichment p-values. The table report that the RobKMR method has discovered different significant genes. To show the performance of these genes, we also consider $7$  genes for an extensive set of functional association data with the GeneMANIA. Figure~\ref{fig:mg} shows the  genes related networks to the  input genes COG3, SMTN, DRGX, MTMD5, CSMD3, DKK1 and  FSTKD2 (at p-value $\leq 0.00001$). This figure suspects that the selected genes have strong physical interactions,  co-expression,  predicted, co-localization, genetic interactions, pathway, and shared protein domains networks. Thus, the proposed robust methods (RobKMR) can find undiscovered genes in addition to significant gene triplets.

\begin{table}
\caption{The GeneHancer identifier, GeneHancer score, gene association score, total score, major-related diseases, and PubMed database (PMID)  of  6 selected   genes}
\begin{center}
\label{tbl:ostdatagene}
\scalebox{0.6}[0.6]{
 \begin{tabular}{l|c|c|c|c|c|c} \hline
\rm{Gene}  & \rm{ GeneHancer} & \rm{GeneHance}r& \rm{Association} &  \rm{Total}& \rm{Transcription Factor
}  &  \rm{Related} \\ 
 \rm{ID} &  \rm{ID} & \rm{Score} &  \rm{Score}&  \rm{ Score} & \rm{Binding Sites (TFs)}  & \rm{ Disease} \\ \hline
\rm{DKK1} & \rm{GH10J052312}& $2.0$&$5256.8$& $520.05$& $140$ & \rm{ bone formation and bone disease, Osteoporosis, cancer (Leukemia) and Alzheimer disease} \\ \hline
\rm {MTND5} & \rm{ GH21J043724}& $2.4$&$9.7$& $23.76$& $340$ & \rm{ brain and muscles} \\ \hline
\rm {COG3} & \rm{GH20J056524}& $0.4$&$500.7$& $221.23$& $(2)$: $25367360$; $25130324$& \rm {Bone marrow;  thyroid} \\ \hline
\rm {FASTKD2} & \rm{GGH02J206764}& $1.7$&$257.7$& $447.37$& $179$& \rm{Mitochondrial disease} \\ \hline
\rm {SMTN} & \rm{GH22J031077}& $2$&$271.3$& $549.79$& $184$& \rm{actin binding and structural constituent of muscle, Glomuvenous Malformations, and Viral Gastritis} \\ \hline
\rm {CSMD3} & \rm{
GH08J113434}& $1.7$&$254.5$& $424.63$&$38 $& \rm{ Familial Adult Myoclonic Epileps and Trichorhinophalangeal Syndrome} \\ \hline
\rm {DRGX} & \rm{
GH10J049396}& $1$&$5258.5$& $270.84$& $6$& \rm{Chromosome 16P12.2-P11.2 Deletion Syndrome, 7.1- To 8.7-Mb} \\ \hline
\end{tabular}}
\end{center}
 \end{table}

\begin{table}
\begin{center}
 \caption {Network analysis of selected $20$  unique genes  by the proposed method at p-value $\leq$ $0.001$}
\scalebox{0.6}[.8]{
\label{tbl:norm}
 \begin{tabular}{l|cccccc} \hline
Gene& Number of & Average node &  Average  local &
Expected number & Interaction enrichment \tabularnewline
&  edges& degree&  clustering coefficient&
 of edges&  p-value\tabularnewline \hline
 &  & & Genome&& \tabularnewline \hline
CLCN7&$29$&$5.27$&$0.84$&$10$&$1.06e-06$\tabularnewline 
DKK1&$55$&$	10.00$&$	1$&$	11$&$	1.0e-16$\tabularnewline 
INSIG2&$30$&$5.45$&$	0.775$&$	11$&$	7.37e-07$\tabularnewline 
JAG1&$40$&$	7.27$&$	0.882$&$	11$&$	3.42e-11$\tabularnewline 
LIN7C&$32$&$5.82$&$	0.874$&$	10$&$	4.25e-08$\tabularnewline 
MBL2&$55$&$	10.00$&$	1$&  $	10$&$1.0e-16$\tabularnewline 
NTAN1&$24$&$4.36$&$	0.88$&$	10$&$	0.000179$\tabularnewline 
RGCC&$40$&$	7.27$&$	0.856$&$	25$&$	0.003$\tabularnewline 
SIDT1&$	19$&$3.45$&$	0.891$&$	10$&$	0.0102$\tabularnewline 
SLC25A13&$	54$&$9.82$&$	0.982$&$	11$&$	1.0e-16$\tabularnewline 
WNT3&$45$&$	8.18$&$	8.18$&$	10$&$	2.66e-15$\tabularnewline \hline
 &  & &  Epigenome && \tabularnewline \hline
COG3&$		53$&$	9.64$&$	0.968$&$	11$&$	1.0e-16$\tabularnewline 
DNER&$	55$&$	10.00$&$	1$&$	11$&$	1.0e-16$\tabularnewline 
DNMBP&$		25$&$	4.55$&$	0.838$&$	11$&$	0.000262$\tabularnewline 
RBM38&$		22$&$	4.00$&$	0.77$&$	13$&$	0.0205$\tabularnewline 
SMTN&$29$&$	5.27$&$	0.913$&$	11$&$	5.21e-06$\tabularnewline \hline
 &  & &  Transcriptome&& \tabularnewline \hline
DRGX&$		16$&$	2.91$&$	0.921$&$	10$&$	0.541$\tabularnewline 
SOX1&$		49$&$	8.91$&$	0.901$&$	12$&$	7.77e-16$\tabularnewline 
UCKL1&$		40$&$	7.27$&$	0.872$&$ 11	$&$5.05e-12$\tabularnewline 
\end{tabular}
}
\end{center}
 \end{table}

\begin{figure}
\begin{center}
\includegraphics[width=8cm, height=8cm]{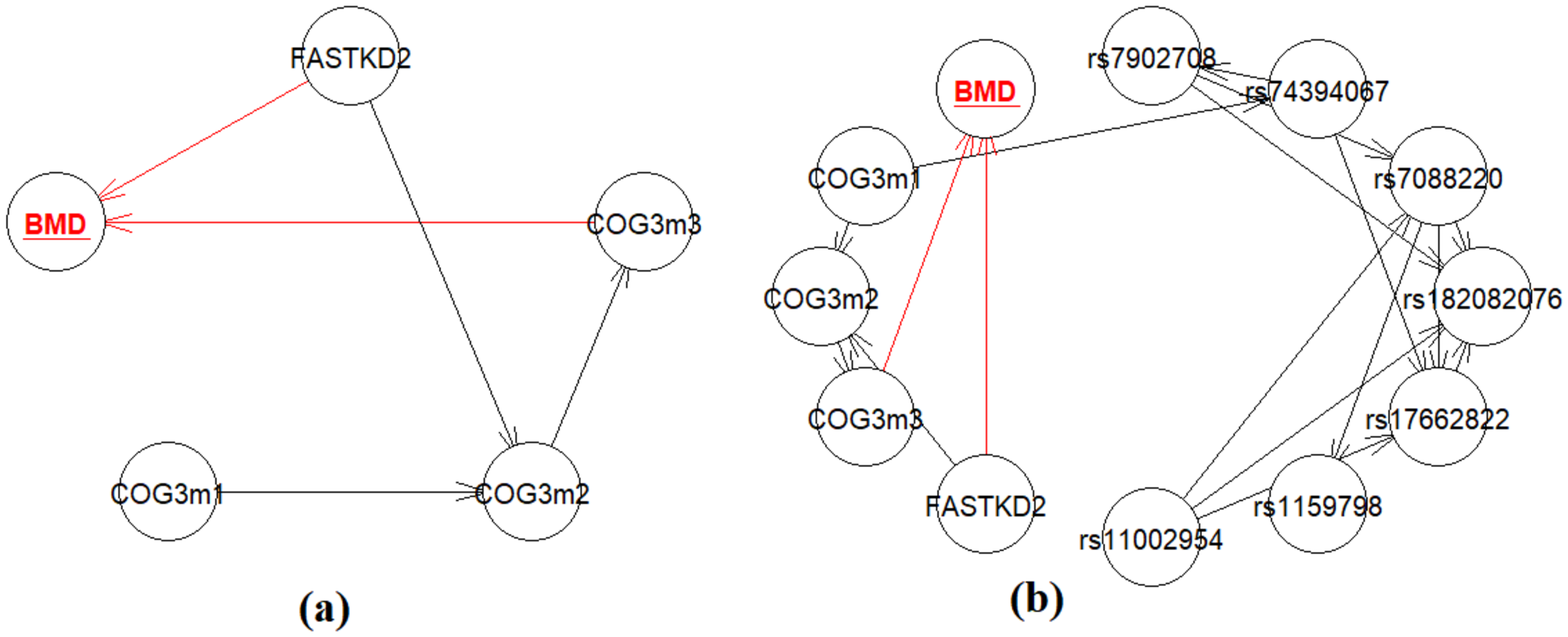}
 \caption{The causal relationship of BMD  drives with the epigenome only (a)  and the epigenome and  genome (b) datasets. Two genes of the epigenome,  FASTKD2 (one methylation profile) and  COG3 (three methylation profiles) and one gene of genome (7 SNPs)  data, are causal related to BMD.} 
\label{fig:cr2d}
\end{center}
\end{figure}

We infer a causal relationship to explore further whether the selected genes will have unique associations with the  BMD. Hill climbing, a greedy search engine, obtain optimal solutions for convex problems instead of local optima for a causal relationship. This approach selects the best successor node under the evaluation function and commits the search \cite{Taskinen-06}. We apply a Hill-Climbing approach among genome and epigenome datasets  (the select genes at p-values $\leq 0.00001$). The causal relationship of BMD with (a) epigenome and (b)  genome and epigenome datasets are illustrated in Figure~\ref{fig:cr2d}. We observed that these two genes,  FASTKD2 ( with one methylation profile) and  COG3 (with three methylation profiles), are directly related to BMD.  Thus, This observation concluded that the selected biomarkers may have significant impact on BMD but are not general.

\subsection{ Drug repositioning for  osteoporosis}
We conducted a literature review of  OP disease for exploring candidate drugs that have host transcriptome-guided 35 meta-drug agents. Thus we considered 35 drug agents to explore candidate drugs by molecular docking with our selected $3$ genes (proposed receptors proteins). To offer in-silico validated efficient candidate drugs for the treatment against OP, we employed a molecular docking study of our proposed receptor proteins with the drug agents. 

\par
\begin{figure}
\begin{center}
\includegraphics[width=8cm, height=8cm]{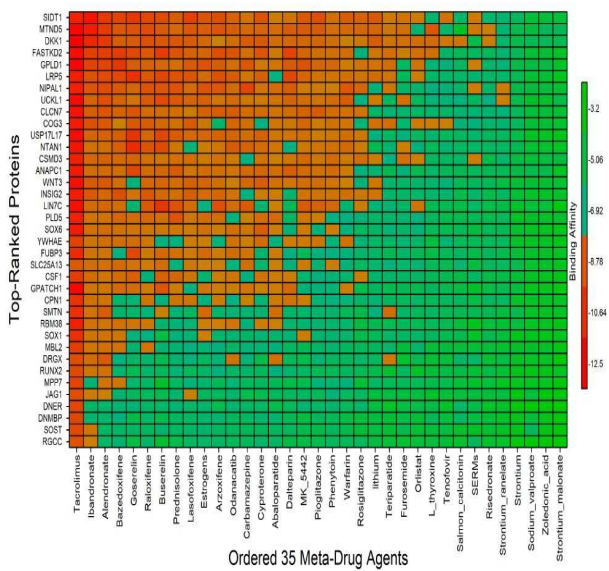}
 \caption{Molecular docking results computed with autodock vina. Image of binding affinities based on the 35 ordered drugs in the X-axis and ordered $37$ proteins (selected genes by the RobKMR at p-values $\leq 0.001$) in the Y-axis. The strong and weak binding affinities between proteins and drugs are denoted by red and green colors, respectively.} 
\label{fig:DD1}
\end{center}
\end{figure}

 We downloaded the 3D structure of the 20 proteins (MBL2, DNMBP, YWHAE, RBM38, SMTN, CPN1, JAG1, LIN7C, MPP7, DKK1, CSF1, SLC25A13, ANAPC1, MTND5, NTAN1, WNT3, SOST, INSIG2, RUNX2, CLCN7) from Protein Data Bank (PDB) with source codes 1hup, 1ug1, 2br9, 2cqd, 2d87, 2nsm, 2vj2, 3lra, 3o46, 3s2k, 3uez, 4p5w, 4ui9, 5xtc, 6a0e, 6ahy, 6l6r, 6m49, 6vg8, 7bxu, respectively \cite{Berman-00}. On the other hand, the 3D structure of 17 proteins (DRGX, USP17L17, SOX1, LRP5, SOX6, GPLD1, NIPAL1, PLD5, DNER, FUBP3, COG3, GPATCH1, RGCC, UCKL1, SIDT1, FASTKD2, CSMD3) were downloaded from AlphaFold source using UniProt  ID of A6NNA5, D6RBQ6, O00570, O75197, P35712, P80108, Q6NVV3, Q8N7P1, Q8NFT8, Q96I24, Q96JB2, Q9BRR8, Q9H4X1, Q9NWZ5, Q9NXL6, Q9NYY8, Q7Z407 \cite{UniProt-19}. We downloaded the 3D structures of 35 drugs from the PubChem database \cite{Kim-19}. Then molecular docking was carried out between $37$ proteins and $35$ meta-drug agents to calculate the binding affinity scores (kcal/mol) for each pair of proteins and drugs. Then we organized  the proteins in descending order of row sums of the binding affinity matrix and drug agents according to the column sums of the scoring matrix to select a few drug agents as the candidate drugs. Figure~ \ref{fig:DD1} presents the binding affinity matrix. Thus we set eight top-ranked drug agents (Tacrolimus, Ibandronate, Alendronate, Bazedoxifene, Goserelin, Raloxifene, Buserelin, Prednisolone) as candidate drugs with average binding affinity scores $-7.5$ kcal/mol $\leq$ against the $37$ proteins.

\par
\begin{figure}[ht]
\begin{center}
\includegraphics[width=8cm, height=8cm]{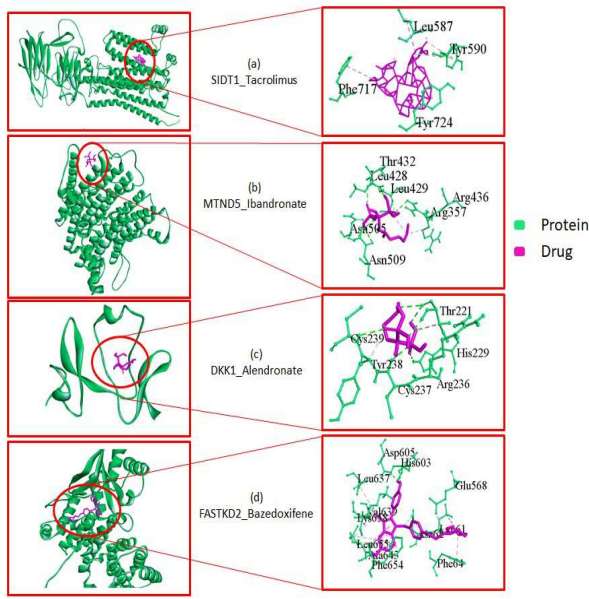}
 \caption{ The 3D structure (left)  and 2D Schematic diagram (right)  of hub protein of the top four potential targets (DKK1,  MTND5, FASTKD2, and SIDT1) and top four lead drugs. We calculated drug interacting amino acid within 4 amstrong. Lead four drugs, Tacrolimus, Ibandronate, Alendronate, and Bazedoxifene, are selected by investigating the binding affinity score.} 
\label{fig:DD2}
\end{center}
\end{figure}

 The docked complexes of the top three virtual hits from AutoDock-Vina docking are further considered for protein-ligand interaction profiling. As shown in Figure~ \ref{fig:DD2}(a), the SIDT1\_Tacrolimus complex showed one hydrogen bond with Tyr590 residues. Although the ligand formed significant (key)  hydrophobic interactions with Leu587, Tyr590, Phe717 residues, and Tyr724 deposition showed additional electrostatic interactions with the drug. On the other hand, MTND5\_Ibandronate (Figure~ \ref{fig:DD2}(b)) complex showed five hydrogen bonds with Leu429, Thr432, Arg436, Asn505, Asn509 residues and the significant hydrophobic interactions with Arg357, Leu429residues. In the case of the DKK1\_Alendronate complex, Alendronate formed six hydrogen bonds with Thr221, His229, Arg236, Cys237, Tyr238, Cys239 residues (see Figure~ \ref{fig:DD2}(b)). As shown in  Figure~ \ref{fig:DD2}(b), the FASTKD2\_Bazedoxifene complex showed five hydrogen bonds with Leu61, Asn62, Glu568, His603, Asp605 residues, and the significant hydrophobic interactions with Phe64, Leu637, Val639, Ala643, Phe654, Leu655, Lys658 residues.

\section{Concluding remarks}
\label{Sec:cont}
This paper developed a novel robust approach to identify inter-related risk factors in multi-omics data for drug repurposing in OP. Benchmarking experiments based on simulation and real datasets analysis have shown that our proposed approach provides a competitive performance compared with the existing ones to derive a statistic for testing the inter-related risk factors. The RobKMR method's power, biological validations, and drug repurposing are further demonstrated by its application to synthesized and a real multi-omics dataset in osteoporosis studies. 
\par Simulation studies show that the test's false positive rate and the power analysis for higher-order composite effect are mitigated by fixing the nominal p-value threshold along with other state-of- the-art-methods. The ROC curves also present the power gain by the   RobKMR and stKMA methods in all scenarios. 
\par The real datasets analysis confirmed that the RobKMR could select the inter-related risk factors of OP for drug repurposing. The network analysis shows that the selected genes of each omics also have significantly more interactions than expected. The gene's function is collaborative and biologically relevant to OP. We observe that the selected genes are directly related to the  BMD by a causal analysis. Additionally,  the top four genes DKK1,  MTND5, FASTKD2  (at p-value $\leq$ 0.00001), and SIDT1 (at p-value  $\leq$  0.001) identify  four lead drugs: Tacrolimus, Ibandronate, Alendronate, and Bazedoxifene  from all $30$ experimented candidates for drug repurposing in OP.
\par While our proposed method can identify stable biomarkers (DKK1,  MTND5, FASTKD2, and SIDT1)  for OP studies, we acknowledge that further investigation of these four genes is essential to improve OP studies for its drug discovery. Further,  the proposed approach can be applied to any disease model where multi-omics datasets are available.




\subsection*{Acknowledgments}
 This work is benefited by the support of U19AG05537301 and
R01AR069055.

\bibliographystyle{plain}
\bibliography{document}

\end{document}